\documentclass{article}

\PassOptionsToPackage{numbers}{natbib}
\usepackage[preprint]{neurips_data_2021}

\usepackage[utf8]{inputenc} % allow utf-8 input
\usepackage[T1]{fontenc}    % use 8-bit T1 fonts
\usepackage{hyperref}       % hyperlinks
\usepackage{url}            % simple URL typesetting
\usepackage{booktabs}       % professional-quality tables
\usepackage{amsfonts}       % blackboard math symbols
\usepackage{nicefrac}       % compact symbols for 1/2, etc.
\usepackage{microtype}      % microtypography
\usepackage{xcolor}         % colors
\usepackage{graphicx}
\usepackage{subcaption}
\usepackage{amsmath}
\usepackage{url}
\usepackage{hyperref}
\usepackage{wrapfig}

\newlength{\figureWidth}
\title{The VVAD-LRS3 Dataset for Visual Voice Activity Detection}

\author{%
    \parbox{\linewidth}{
        \centering
        Adrian Lubitz$^1$ \quad Frank Kirchner$^{1,2}$ \quad Matias Valdenegro-Toro$^2$\\
    }\\
    ~\\
    \parbox{\linewidth}{
        \centering
        $^1$ Department of Computer Science, University of Bremen, Bremen, Germany.\\
        $^2$ Robotics Innovation Center, German Research Center for Artificial Intelligence, Bremen, Germany.\\
        ~\\
        \texttt{alubitz@uni-bremen.de}, \texttt{frank.kirchner@dfki.de}, \texttt{matias.valdenegro@dfki.de}
    }
}

\begin{document}

\maketitle

\begin{abstract}
    Robots are becoming everyday devices, increasing their interaction with humans.
To make human-machine interaction more natural, cognitive features like
Visual Voice Activity Detection (VVAD), which can detect whether a
person is speaking or not, given visual input of a camera, need to be implemented.
Neural networks are state of the art for tasks in Image Processing, Time Series Prediction, Natural Language Processing and other domains.
Those Networks require large quantities of labeled data.
Currently there are not many datasets for the task of VVAD. In this work we created a large scale dataset called the
VVAD-LRS3 dataset, derived by automatic annotations from the LRS3 dataset.
The VVAD-LRS3 dataset contains over 44K samples, over three times the next competitive dataset (WildVVAD).
We evaluate different baselines on four kinds of features:
facial and lip images, and facial and lip landmark features.
With a Convolutional Neural Network Long Short Term Memory (CNN LSTM) on facial images an accuracy of 92\% was reached on the test set.
A study with humans showed that they reach an accuracy of 87.93\% on the test set.
\end{abstract}

\section{INTRODUCTION}
Technology is integrating more and more into the life of the modern man. A very important
question is how are people interacting with technology.
The human brain does not react emotionally to artificial objects like computers and
mobile phones. However, the human brain reacts strongly to human appearances like
shape of the human body or faces \cite{shu533}. Therefore humanoid robots are the most
natural way for human-machine interaction, because of the human-like appearance.
This hypothesis is strongly supported by HRI Research \cite{kanda2017human, PascWhen2013, doi:10.1142/S0219843605000582, 1242146}.
They see Social Robots as a part of the future society. % as shown in figure \ref{fig:socialRobotsFuture}.
\cite{kanda2017human} also defines the following three issues which
need to be solved to bring social robots effectively and safely to the everyday life:
\begin{itemize}
    \item [a.] Sensor network for tracking robots and people
    \item [b.] Development of humanoids that can work in the daily environment.
    \item [c.] Development of functions for interactions with people.
\end{itemize}
This paper is located in the field \texttt{c},
as we propose a large scale dataset to train models for the task of
Visual Voice Activity Detection (VVAD)
which detects whether a person is speaking to a robot or not,
given the visual input of the robot's camera.

\makeatletter%
\if@twocolumn%
    \figureWidth=\columnwidth
\else% \@twocolumnfalse
    \figureWidth=0.45\columnwidth
\fi
\makeatother

% \makeatletter%
% \if@twocolumn%
%     \begin{figure}
%         \centering
%         \includegraphics[width=\figureWidth]{images/socialRobotsFuture.png}
%         \caption{Social Robots in the future society \cite{kanda2017human}}
%         \label{fig:socialRobotsFuture}
%     \end{figure}
% \else% \@twocolumnfalse
%     \begin{wrapfigure}{r}{0.5\textwidth}
%         \centering
%         \includegraphics[width=\figureWidth]{images/socialRobotsFuture.png}
%         \caption{Social Robots in the future society \cite{kanda2017human}}
%         \label{fig:socialRobotsFuture}
%     \end{wrapfigure}
% \fi
% \makeatother

VVAD is an important cognitive feature in a Human-Robot Interaction(HRI).
As we want Robots to integrate seamlessly into our society, Human-Robot Interaction needs to be as close as possible to Human-Human Interaction (HHI).
VVAD can be used for speaker detection in the case where multiple people are in the robot's field of view.
Furthermore it can be usefull to detect directed speech in noisy environments.

In this paper we present a new benchmark for the VVAD task, produced from the LRS3 dataset \cite{Chung18} which contains TED Talks,
and by using the provided textual transcripts, we can extracts parts of the TED Talk video in order to generate
positive/negative video samples for the VVAD task. Our dataset contains 37.6K training and 6.6K validation samples,
making it the largest VVAD dataset currently. The dataset will be publicly available on the internet.
We provide baseline models using commonly used neural network architectures.
In an experimental setup with a CNN LSTM an accuracy of 92\% was reached on the test set.
A study with humans showed that humans reach an accuracy of 87.93\% on the test set.

\textbf{This paper contributes a large scale dataset and a simple approach on how to use it for VVAD.}

\section{RELATED WORK}\label{sec:relatedWork}
The classic approach to solve VVAD is to detect lip motion.
This approach is taken by F. Luthon and  M. Liévin in \cite{Luthon1998}. They try to model the motion of the mouth in a sequence of color images with Markov Random Fields. For the lip detection they analyze the images in the \emph{HIS (Hue, Intensity, Saturation)} color space, with extracting \emph{close-to-red-hue prevailing regions} this leads to a robust lightening independent lip detection.
A different approach was taken by Spyridon Siatras, Nikos Nikolaidis, and Ioannis Pitas in \cite{Siatras2006}. They try to convert the problem of lip motion detection into a signal detection problem. They measure the intensity of pixels of the mouth region and classify with a threshold, since they argue that frames with an open mouth have a essentially higher number of pixels with low intensity.
In \cite{Bendris2010} Meriem Bendris, Delphine Charlet and Gérard Chollet propose a method,
which measures the probability of voice activity with the optical flow of pixels in the mouth region.
In \cite{Bendris2010} the drawback of lip motion detection based approaches is already discussed.
As shown in Figure \ref{fig:errorDetection}, what makes the problem difficult is that people move their lips from time to time although they are not speaking.

\makeatletter%
\if@twocolumn%
    \figureWidth=\columnwidth
\else% \@twocolumnfalse
    \figureWidth=0.45\columnwidth
\fi
\makeatother

\makeatletter%
\if@twocolumn%
    \begin{figure}
        \centering
        \includegraphics[width=\figureWidth]{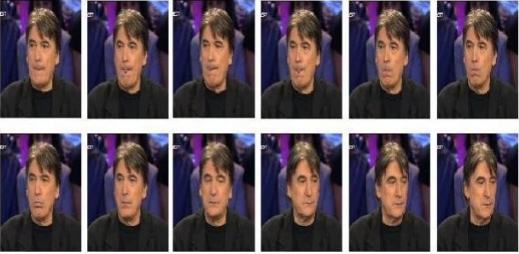}
        \caption{Example of error detection - Person is classified as
            having a mouth activity, however does not speak \cite{Bendris2010}}
        \label{fig:errorDetection}
    \end{figure}
\else% \@twocolumnfalse
    \begin{wrapfigure}{r}{0.5\textwidth}
        \centering
        \includegraphics[width=\figureWidth]{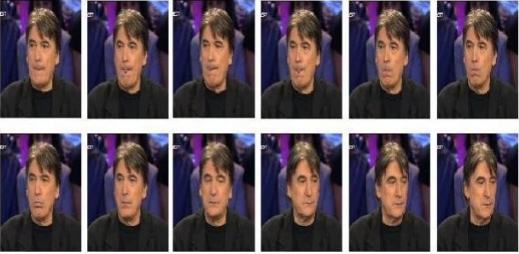}
        \caption{Example of error detection - Person is classified as
            having a mouth activity, however does not speak \cite{Bendris2010}}
        \label{fig:errorDetection}
    \end{wrapfigure}
\fi
\makeatother

This issue is tackled by Foteini Patrona, Alexandros Iosifidis et al. \cite{Patrona2016}.
They use a Space Time Interest Point (STIP) or the Dense Trajectory-
based facial video representation to train a Single Hidden Layer Feedforward Neural Network.
The features are generated from the CUAVE dataset \cite{Patterson2002}.
This erases the implicit assumption (of the approaches above) that lip motion equals voice activity.
A more robust approach, which uses Centroid Distance Features of normalized lip shape to train a LSTM Recurrent Neural Network is proposed by Zaw Htet Aung and Panrasee Ritthipravat in \cite{Aung2015}. This method shows a classification accuracy up to 98\% on a relatively small dataset.
In conclusion all of the mentioned methods use some kind of face detection and some also use mechanics to track the face. This is needed if there is more than one face in the image.
From the facial images features are created in different ways. From that point the approaches divide into two branches.
The first and naive approach is to assume that lip motion equals speech. This is obviously not always the case, which is why the later approaches do not rely on this hypothesis.
The latter approach uses learning algorithms to learn the real mapping between facial images and the speech/no speech. This approach is strongly relying on a balanced dataset to learn a good performing model.

While datasets like the LRS3 or CUAVE \cite{Patterson2002} provide a good fit for lipreading they lack the negative class for VVAD. There seems to be not many datasets for the VVAD task.
The only competitive state of the art dataset for VVAD that we found was the WildVVAD \cite{guy2020learning}.
WildVVAD is not only 3 times smaller than the VVAD-LRS3 it is also more prone false positive and false negative
because of the loose assumption that detected voice activity and a single face in the video equals a \emph{speaking} sample
and every detected face in a video sequence without voice activity is a \emph{not speaking} sample.
Furthermore the source WildVVAD is drawn from makes it less diverse.
Table \ref{tb:datasets} shows a comparison of state of the art datasets. The VVAD-LRS3 that we propose in this paper is $\sim 3 \times$ larger than WildVVAD.

\begin{table}[t]
    \caption{Overview of state of the art datasets for VVAD}
    \label{tb:datasets}
    \centering
    \begin{tabular}{llll}
        \toprule
        \textbf{Dataset}                & \textbf{Samples} & \textbf{Diversity} & \textbf{Pos/Neg Ratio} \\
        \midrule
        VVAD-LRS3 (this work)           & 44,489           & Very high          & 1-to-1                 \\
        WildVVAD \cite{guy2020learning} & 13,000           & High               & 1-to-1                 \\
        LRS3 \cite{Chung18}             & $>$100,000       & High               & 1-to-0                 \\
        CUAVE \cite{Patterson2002}      & $\sim$7,000      & Low                & 1-to-0                 \\
        \bottomrule
    \end{tabular}

\end{table}

\section{DATASET CAPTURE}
\label{sec:data}

To create the large scale VVAD dataset we took the
Lip Reading Sentences 3 (LRS3) Dataset introduced by Afouras et al. in \cite{Chung18} as a basis.
The LRS3 is a dataset designed for visual speech recognition and is created from videos of 5594 TED and TEDx talks.
It provides more than 400 hours video material of natural speech.
The LRS3 dataset provides videos along with metadata about the face position and a speech transcript. In the LRS3 metadata files the following fields are important for the transformation to the VVAD dataset:
\paragraph{Text} contains the text for one sample. The length of the text or respectively the sample is defined by length of the scene.
That means one sample can get as long as the face is present in the video.
\paragraph{Ref} is the reference to the corresponding YouTube video. The value of this field needs to be appended to \texttt{https://www.youtube.com/watch?v=}
\paragraph{FRAME} corresponds to the face bounding box for every frame, where \texttt{FRAME} is the frame number, \texttt{X} and \texttt{Y} is the position of the bounding box in the video and \texttt{W} and \texttt{H} are the width and height of the bounding box respectively. It is to mention that for the frame number a frame rate of 25 fps is assumed and the values for \texttt{X}, \texttt{Y}, \texttt{W} and \texttt{H} are a percentage indication of the width and height of the video.
\paragraph{WORD} maps a timing to every said word. Here \texttt{START} and \texttt{END} indicate the start and end of the word in seconds respectively.
It is to mention that the time is in respect to the start of the sample given by the first frame and not to the start of the whole video.

The LRS3 dataset comes with a low bias towards specific ethnic groups, because TED and TEDx talks are international and talks are held by men and women as well as small children.
It also comes with the advantage that it depicts a large variety of people because the likelihood of talking in multiple TED or TEDx talks is rather small.
This is a big advantage over the LRS2 and LRW dataset that are extracted from regular TV shows, which brings the risk of overfitting to a specific persons.
LRS3 makes learning more robust in that sense.
Since natural speech in front of an audience includes pauses for applause and
means to structure and control a speech as described in \cite{Nikitina2011}, the LRS3 dataset provides
\emph{speaking} and \emph{not speaking} phases.

\begin{table}[t]
    \caption{Number of samples for training, validation and test splits of the VVAD-LRS3 dataset}
    \label{tb:samples}
    \centering
    \begin{tabular}{lll}
        \toprule
        \textbf{Training Set} & \textbf{Validation Set} & \textbf{Test Set} \\
        \midrule
        37646 Samples         & 6643 Samples            & 200 Samples       \\
        \bottomrule
    \end{tabular}

\end{table}

\begin{figure}
    \centering
    \parbox{\figureWidth}{
        \begin{subfigure}{.5\figureWidth}
            \centering
            \captionsetup{width=.8\linewidth}
            \includegraphics[width=\linewidth]{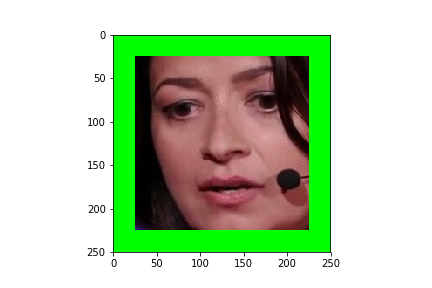}
            \caption{Face Images feature}
        \end{subfigure}%
        \begin{subfigure}{.5\figureWidth}
            \centering
            \captionsetup{width=.8\linewidth}
            \includegraphics[width=\linewidth]{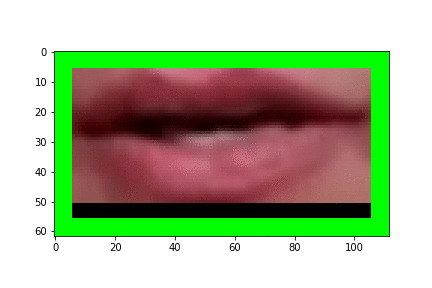}
            \caption{Lip Images feature}
        \end{subfigure}
    }
    \parbox{\figureWidth}{
        \begin{subfigure}{.5\figureWidth}
            \centering
            \captionsetup{width=.8\linewidth}
            \includegraphics[width=\linewidth]{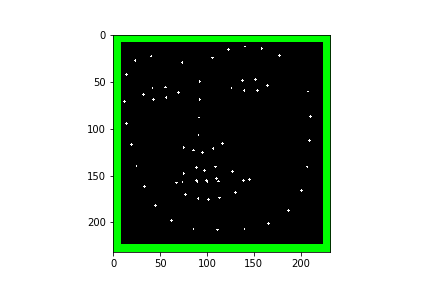}
            \caption{Face Features feature}
        \end{subfigure}%
        \begin{subfigure}{.5\figureWidth}
            \centering
            \captionsetup{width=.8\linewidth}
            \includegraphics[width=\linewidth]{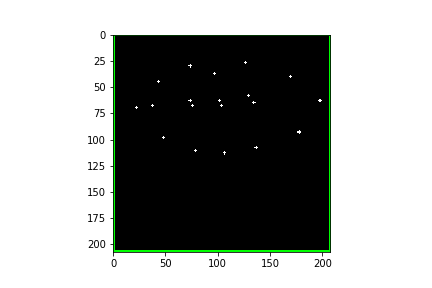}
            \caption{Lip Features feature}
        \end{subfigure}%
    }
    \caption{Visualization of one frame of different features.}
    \label{fig:flavors}
\end{figure}

To transform LRS3 samples to VVAD ones the given text files are analyzed for these \emph{speaking} and \emph{not speaking} phases.
In \cite{BrigPaus1994} Brigitte Zellner
shows that pauses occur in natural speech and explicitly in speech in front of an audience.
This leads to two constants we need to define in the context of pauses.
The first is \texttt{maxPauseLength} which defines the maximal length of a pause which is still considered to be an inter speech pause.
In consideration of the different types of pauses mentioned in \cite{BrigPaus1994} \texttt{maxPauseLength} is set to 1 s.
The second constant is \texttt{sampleLength} which defines the length of a sample.
In other words this defines how long a pause should be to be considered as a negative (\emph{not speaking}) sample or how long a speech phase needs to be to be considered a positive (\emph{speaking}) sample.
It shows that most of the pauses have a length between 1.5 s and 2.5 s, therefore \texttt{sampleLength} is set to 1.5 s to get the most out of the LRS3 dataset.
The extraction of positive and negative samples for the VVAD starts only on textual basis.
Theoretically the whole extraction of the data could work on this basis but the given bounding boxes where very poor.
To overcome this problem face detection and tracking was remade using dlib's \cite{dlib09} correlation filter based tracker and face detector. We provide for different kinds of features derived from the tracked face and facial features:
\begin{itemize}
    \item \textbf{Face Images}  The whole image resized and zero padded to a specific size.
    \item \textbf{Lip Images} An image of only the lips resized and zero padded to a specific size.
    \item \textbf{Face Features}  All 68 facial landmarks extracted with dlib's facial landmark detector
    \item \textbf{Lip Features} All facial landmarks concerning the lips extracted with dlib's facial landmark detector
\end{itemize}
For the face images the input image only needs to be resized and zero padded to a given size.
As depicted in Figure \ref{fig:flavors} the predictor extracts facial shape given by 68 landmarks, while 20 of these landmarks describe the lips.
The predictor is trained on the ibug 300-W face landmark dataset \footnote{Available at \href{https://ibug.doc.ic.ac.uk/resources/facial-point-annotations/}{https://ibug.doc.ic.ac.uk/resources/facial-point-annotations/}}.
For the \emph{lip images} the minimal values in x- and y-direction are taken as the upper left corner of the lip image, while the lower right corner is defined by the maximal values in x- and y-direction.
The \emph{face features} are taken directly from the landmarks given from dlib's facial landmark detector.
For the \emph{lip features} only landmark 49 to landmark 68 are taken into account, because they fully describe the lip shape as seen in Figure \ref{fig:flavors}.
It is to mention that it is useful to normalize the features for \emph{face features} and \emph{lip features} when applied to a learning algorithm.

\makeatletter%
\if@twocolumn%
    \figureWidth=\columnwidth
\else% \@twocolumnfalse
    \figureWidth=0.65\columnwidth
\fi
\makeatother

\begin{figure*}[t]
    \begin{subfigure}{\textwidth}
        \includegraphics[width=0.09\textwidth]{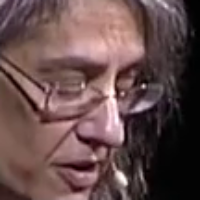}
        \includegraphics[width=0.09\textwidth]{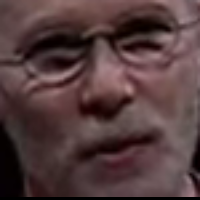}
        \includegraphics[width=0.09\textwidth]{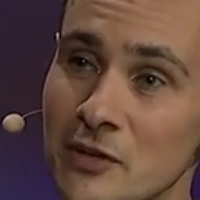}
        \includegraphics[width=0.09\textwidth]{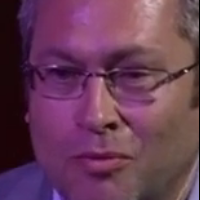}
        \includegraphics[width=0.09\textwidth]{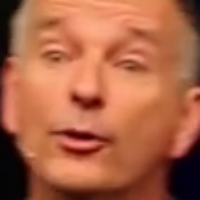}
        \includegraphics[width=0.09\textwidth]{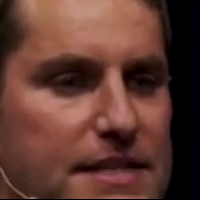}
        \includegraphics[width=0.09\textwidth]{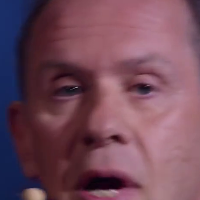}
        \includegraphics[width=0.09\textwidth]{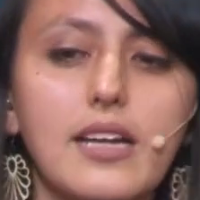}
        \includegraphics[width=0.09\textwidth]{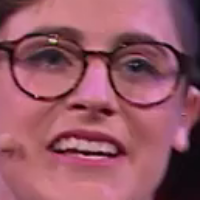}
        \includegraphics[width=0.09\textwidth]{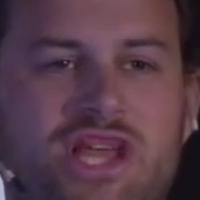}
        \caption{Positive Examples}
    \end{subfigure}
    \begin{subfigure}{\textwidth}
        \includegraphics[width=0.09\textwidth]{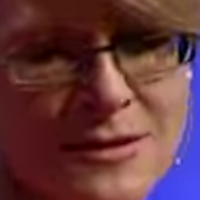}
        \includegraphics[width=0.09\textwidth]{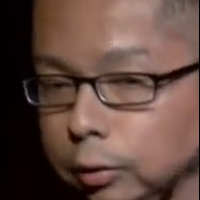}
        \includegraphics[width=0.09\textwidth]{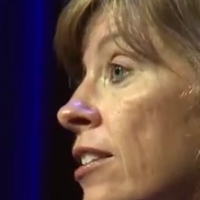}
        \includegraphics[width=0.09\textwidth]{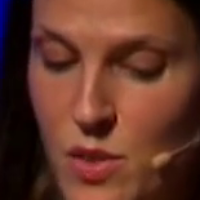}
        \includegraphics[width=0.09\textwidth]{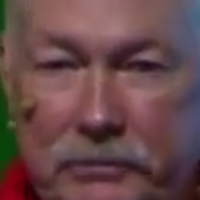}
        \includegraphics[width=0.09\textwidth]{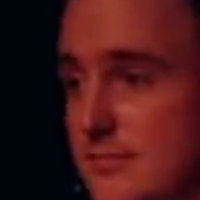}
        \includegraphics[width=0.09\textwidth]{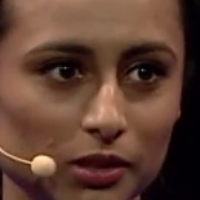}
        \includegraphics[width=0.09\textwidth]{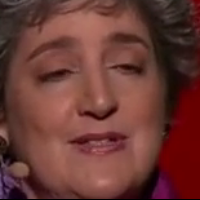}
        \includegraphics[width=0.09\textwidth]{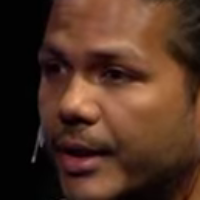}
        \includegraphics[width=0.09\textwidth]{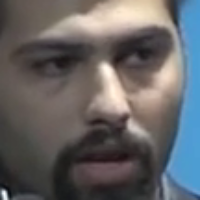}
        \caption{Negative Examples}
    \end{subfigure}

    \caption{Random selection of speaking (positive) and negative (not speaking) samples from the VVAD-LRS3 dataset}
    \label{fig:randomselection}
\end{figure*}

With this approach we could create 22,245 negative (\emph{not speaking}) samples and 22,244 positive (\emph{speaking}) samples which is equal to 18.5 h of learning data in total.
While the theoretical number of positive samples is way higher we were aiming for a balanced dataset and experimental results show that this is sufficient. Table \ref{tb:samples} shows the number of samples on the training, validation and test sets. Figure \ref{fig:randomselection} shows a random selection of 10 positive and negative images from the training set. Table \ref{tb:flavors} shows the dimensionality of one sample for the different features.

\begin{table}[t]
    \caption{Dimensionality for the different features}
    \label{tb:flavors}
    \centering
    \begin{tabular}{llll}
        \toprule
                               & \textbf{Timesteps} & \textbf{Dimensions}       & \textbf{dtype}   \\
        \midrule
        \textbf{Face Images}   & 38                 & $200 \times 200 \times 3$ & \texttt{uint8}   \\
        \textbf{Lip Images}    & 38                 & $100 \times 50 \times 3$  & \texttt{uint8}   \\
        \textbf{Face Features} & 38                 & $68 \times 2$             & \texttt{float64} \\
        \textbf{Lip Features}  & 38                 & $20 \times 2$             & \texttt{float64} \\
        \bottomrule
    \end{tabular}
\end{table}

\begin{figure}
    \centering
    \begin{subfigure}{0.49\textwidth}
        \includegraphics[width=0.90\figureWidth]{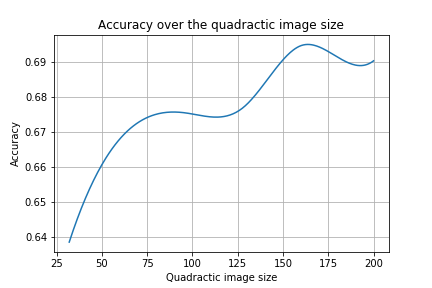}
        \caption{Evaluation of the optimal image size with a single frame}
        \label{fig:AccOverImagesize}
    \end{subfigure}
    \begin{subfigure}{0.49\textwidth}
        \includegraphics[width=0.90\figureWidth]{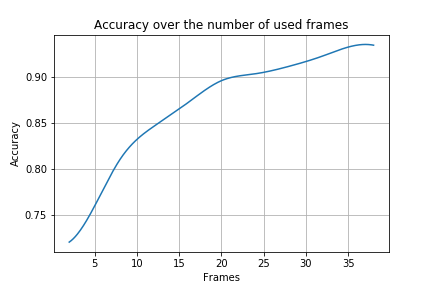}
        \caption{Validation Accuracy for CNN LSTM over the number of timesteps/frames used.}
        \label{fig:AccTimeStepsMobile}
    \end{subfigure}
    \caption{Comparison of performance as image size and number of timesteps/frames is varied on MobileNet.}
\end{figure}

We evaluate two important hyper-parameters of our dataset to examine their relation with learning performance:

\textbf{Image Size}. The optimal image size is evaluated for MobileNets \cite{howard2017mobilenets} using image sizes starting from $32 \times 32$ to the maximal image size of $200 \time 200$ with a step size of 32. Figure \ref{fig:AccOverImagesize} shows that the maximal accuracy in the spatial domain can be reached using a image size of around $160 \times 160$.

\makeatletter%
\if@twocolumn%
    \figureWidth=\columnwidth
\else% \@twocolumnfalse
    \figureWidth=0.6\columnwidth
\fi
\makeatother

\textbf{Number of Frames}. Figure \ref{fig:AccTimeStepsMobile} shows how accuracy improves over the number of frames for a \emph{TimeDistributed} MobileNet on $96 \times 96$ pixel images (limited by available GPU memory). These results show that the VVAD task requires many frames for an accurate prediction and speaking cannot be inferred from a low number of frames.

Taking Figure \ref{fig:AccTimeStepsMobile} and \ref{fig:AccOverImagesize} into account the optimal values for the image size and number of frames are $160 \times 160$ and 36 respectively.

\setcounter{paragraph}{0}
\textbf{Dataset Construction}. To test the dataset with different models we created the following four features that are available directly on our dataset:
\paragraph{Face Images} used for the most sophisticated model.
These Face Images come in a maximal resolution of $200 \times 200$ pixels and with a maximal number of 38 frames.
So the maximum shape of one sample of the \emph{Face Images} feature is $38 \text{ frames} \times 200 \text{ pixels} \times 200 \text{ pixels} \times 3 \text{ channels}$ = 4.56 MB.
Pixel values range between 0 and 255 which can be represented with one byte.
\paragraph{Lip Images} are also used for an end-to-end learning approach but they obviously concentrate on a small subset of the Face Images. Lip Images are RGB images with a maximum of 38 frames but they have a maximal resolution of $100 \times 50$ pixels. This resolves to $38 \text{ frames} \times 100 \text{ pixels} \times 50 \text{ pixels} \times 3 \text{ channels}$ = 0.57 MB.
\paragraph{Face Features} are used for the learning approach which focuses on facial features.

We provides 68 landmarks with a $(x, y)$ position for a single face as depicted in Figure \ref{fig:flavors}.
A single feature is given as float64 (8 bytes), given by $38 \text{ frames} \times 68 \text{ features} \times 2 \text{ dimensions} \times 8 \text{ bytes}$ = 41.4 KB.
\paragraph{Lip Features} are a small subset of Face Features that only take the features of the lips into account.
dlib's facial landmark detector reserves 20 features for the lips as shown in Figure \ref{fig:flavors}.
This results in the size of $38 \text{ frames} \times 20 \text{ features} \times 2 \text{ dimensions} \times 8 \text{ bytes}$ = 12.1 KB for a single sample in the lip features flavor.

\makeatletter%
\if@twocolumn%
    \figureWidth=\columnwidth
\else% \@twocolumnfalse
    \figureWidth=0.6\columnwidth
\fi
\makeatother

\textbf{Test set and Human-Level Accuracy}.
To test the VVAD-LRS3 dataset a human accuracy test was performed. The test is built with a randomly seeded subset of 200 samples that is not part of the train/val splits, and we used 10 persons to produce predictions for this set. The overall human accuracy level was 87.93\%, while the human accuracy level on positive samples is 91.44\%, and the human accuracy level on negative samples is only 84.44\%.

This shows, that the automatic extraction of the negative samples is more prone to errors than the automatic extraction of positive samples. This is due to the purpose of the LRS3 as a lipreading dataset which obviously offers more positive samples than negative samples for a VVAD dataset.

In the human accuracy test some of the samples were labeled incorrectly or at least were classified with the opposite class label. A closer look is taken into four of these samples from the test set. While the samples \texttt{31366} and \texttt{42768} are labeled positive from the automatic transformation from the LRS3 sample to the VVAD sample they were classified as negative by all the subjects in the human accuracy level test.
For the samples \texttt{14679} and \texttt{6178} the opposite is the case.
On further investigation it was seen that sample \texttt{14679} and \texttt{42768} are obviously wrong labeled while sample \texttt{31366} and \texttt{6178} have some special properties that make them perform very bad on the human accuracy level test.
Sample \texttt{31366} has a very quick head movement which makes it very hard to see the very little movements of the mouth to produce speech.
Sample \texttt{6178} shows a person obviously producing sound with his mouth.
But the sound here is no speech but \emph{beat boxing} which is not considered speech in the original LRS3 dataset.
Sample \texttt{6178} and \texttt{31366} are depicted in Figure \ref{fig:6178} and \ref{fig:31366} respectively.

\makeatletter%
\if@twocolumn%
    \figureWidth=\columnwidth
    \begin{figure}
        \centering
        \parbox{\figureWidth}{
            \begin{subfigure}{0.18\linewidth}
                \centering
                \includegraphics[width=\linewidth]{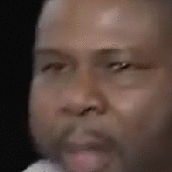}
            \end{subfigure}
            \begin{subfigure}{0.18\linewidth}
                \centering
                \includegraphics[width=\linewidth]{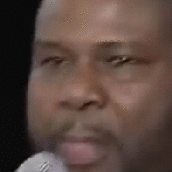}
            \end{subfigure}
            \begin{subfigure}{0.18\linewidth}
                \centering
                \includegraphics[width=\linewidth]{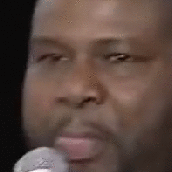}
            \end{subfigure}
            \begin{subfigure}{0.18\linewidth}
                \centering
                \includegraphics[width=\linewidth]{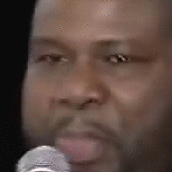}
            \end{subfigure}
            \begin{subfigure}{0.18\linewidth}
                \centering
                \includegraphics[width=\linewidth]{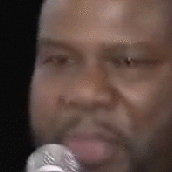}
            \end{subfigure}
        }
        \parbox{\figureWidth}{
            \begin{subfigure}{0.18\linewidth}
                \centering
                \includegraphics[width=\linewidth]{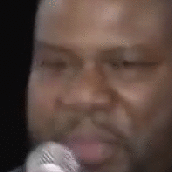}
            \end{subfigure}
            \begin{subfigure}{0.18\linewidth}
                \centering
                \includegraphics[width=\linewidth]{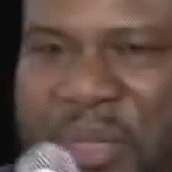}
            \end{subfigure}
            \begin{subfigure}{0.18\linewidth}
                \centering
                \includegraphics[width=\linewidth]{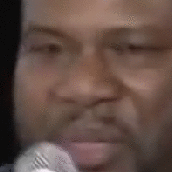}
            \end{subfigure}
            \begin{subfigure}{0.18\linewidth}
                \centering
                \includegraphics[width=\linewidth]{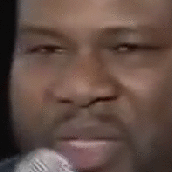}
            \end{subfigure}
            \begin{subfigure}{0.18\linewidth}
                \centering
                \includegraphics[width=\linewidth]{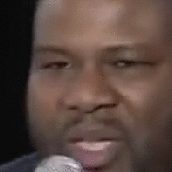}
            \end{subfigure}
        }
        \parbox{\figureWidth}{
            \begin{subfigure}{0.18\linewidth}
                \centering
                \includegraphics[width=\linewidth]{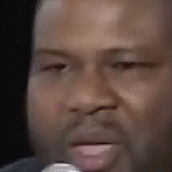}
            \end{subfigure}
            \begin{subfigure}{0.18\linewidth}
                \centering
                \includegraphics[width=\linewidth]{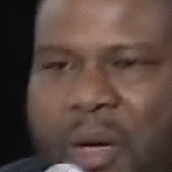}
            \end{subfigure}
            \begin{subfigure}{0.18\linewidth}
                \centering
                \includegraphics[width=\linewidth]{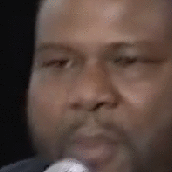}
            \end{subfigure}
            \begin{subfigure}{0.18\linewidth}
                \centering
                \includegraphics[width=\linewidth]{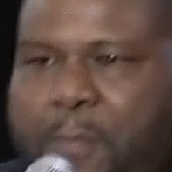}
            \end{subfigure}
            \begin{subfigure}{0.18\linewidth}
                \centering
                \includegraphics[width=\linewidth]{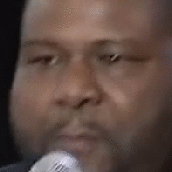}
            \end{subfigure}
        }
        \parbox{\figureWidth}{
            \begin{subfigure}{0.18\linewidth}
                \centering
                \includegraphics[width=\linewidth]{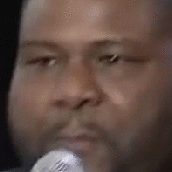}
            \end{subfigure}
            \begin{subfigure}{0.18\linewidth}
                \centering
                \includegraphics[width=\linewidth]{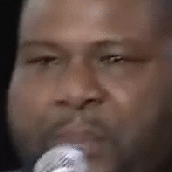}
            \end{subfigure}
            \begin{subfigure}{0.18\linewidth}
                \centering
                \includegraphics[width=\linewidth]{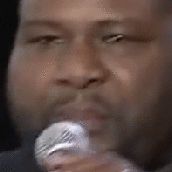}
            \end{subfigure}
            \begin{subfigure}{0.18\linewidth}
                \centering
                \includegraphics[width=\linewidth]{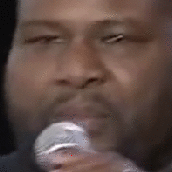}
            \end{subfigure}
            \begin{subfigure}{0.18\linewidth}
                \centering
                \includegraphics[width=\linewidth]{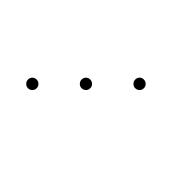}
            \end{subfigure}
        }
        \caption{Sample \texttt{6178} is labeled as a negative (\emph{not speaking}) sample by the automatic transformation from LRS3 to VVAD dataset. On the human accuracy level test 100\% of the subjects classified the sample as positive (\emph{speaking}) sample. Beat boxing is not considered speech in the LRS3 dataset, which causes the wrong label.}
        \label{fig:6178}
    \end{figure}
    \begin{figure}
        \centering
        \parbox{\figureWidth}{
            \begin{subfigure}{0.18\linewidth}
                \centering
                \includegraphics[width=\linewidth]{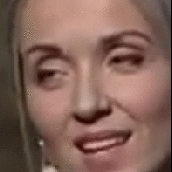}
            \end{subfigure}
            \begin{subfigure}{0.18\linewidth}
                \centering
                \includegraphics[width=\linewidth]{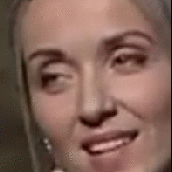}
            \end{subfigure}
            \begin{subfigure}{0.18\linewidth}
                \centering
                \includegraphics[width=\linewidth]{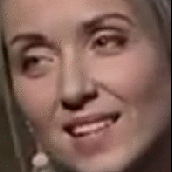}
            \end{subfigure}
            \begin{subfigure}{0.18\linewidth}
                \centering
                \includegraphics[width=\linewidth]{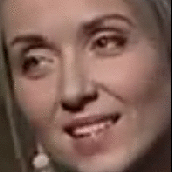}
            \end{subfigure}
            \begin{subfigure}{0.18\linewidth}
                \centering
                \includegraphics[width=\linewidth]{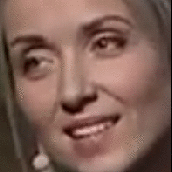}
            \end{subfigure}
        }
        \parbox{\figureWidth}{
            \begin{subfigure}{0.18\linewidth}
                \centering
                \includegraphics[width=\linewidth]{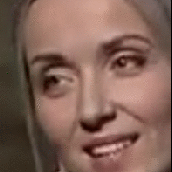}
            \end{subfigure}
            \begin{subfigure}{0.18\linewidth}
                \centering
                \includegraphics[width=\linewidth]{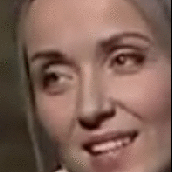}
            \end{subfigure}
            \begin{subfigure}{0.18\linewidth}
                \centering
                \includegraphics[width=\linewidth]{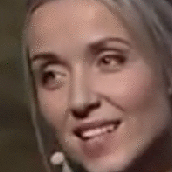}
            \end{subfigure}
            \begin{subfigure}{0.18\linewidth}
                \centering
                \includegraphics[width=\linewidth]{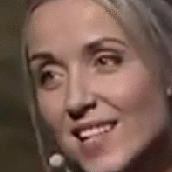}
            \end{subfigure}
            \begin{subfigure}{0.18\linewidth}
                \centering
                \includegraphics[width=\linewidth]{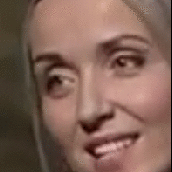}
            \end{subfigure}
        }
        \parbox{\figureWidth}{
            \begin{subfigure}{0.18\linewidth}
                \centering
                \includegraphics[width=\linewidth]{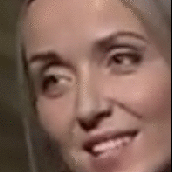}
            \end{subfigure}
            \begin{subfigure}{0.18\linewidth}
                \centering
                \includegraphics[width=\linewidth]{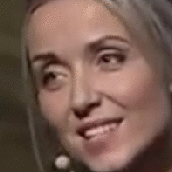}
            \end{subfigure}
            \begin{subfigure}{0.18\linewidth}
                \centering
                \includegraphics[width=\linewidth]{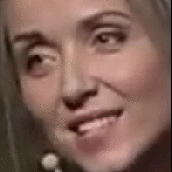}
            \end{subfigure}
            \begin{subfigure}{0.18\linewidth}
                \centering
                \includegraphics[width=\linewidth]{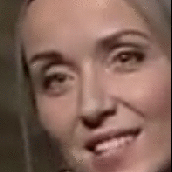}
            \end{subfigure}
            \begin{subfigure}{0.18\linewidth}
                \centering
                \includegraphics[width=\linewidth]{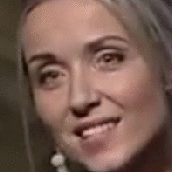}
            \end{subfigure}
        }
        \parbox{\figureWidth}{
            \begin{subfigure}{0.18\linewidth}
                \centering
                \includegraphics[width=\linewidth]{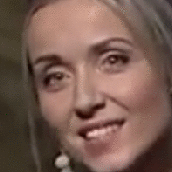}
            \end{subfigure}
            \begin{subfigure}{0.18\linewidth}
                \centering
                \includegraphics[width=\linewidth]{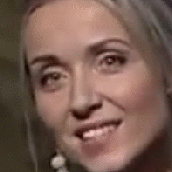}
            \end{subfigure}
            \begin{subfigure}{0.18\linewidth}
                \centering
                \includegraphics[width=\linewidth]{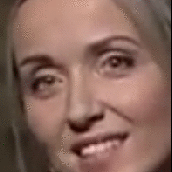}
            \end{subfigure}
            \begin{subfigure}{0.18\linewidth}
                \centering
                \includegraphics[width=\linewidth]{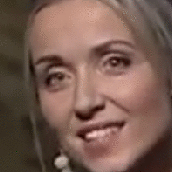}
            \end{subfigure}
            \begin{subfigure}{0.18\linewidth}
                \centering
                \includegraphics[width=\linewidth]{images/dots.jpg}
            \end{subfigure}
        }
        \caption{Sample \texttt{31366} is labeled as a positive (\emph{speaking}) sample by the automatic transformation from LRS3 to VVAD dataset. On the human accuracy level test 100\% of the subjects classified the sample as negative (\emph{not speaking}) sample. The fast movement of the head while producing only a small movement of the lips causes the wrong label.}
        \label{fig:31366}
    \end{figure}
\else% \@twocolumnfalse
    % \figureWidth=0.4\columnwidth
    \begin{figure}
        \centering
        \parbox{\textwidth}{
            \begin{subfigure}{0.09\linewidth}
                \centering
                \includegraphics[width=\linewidth]{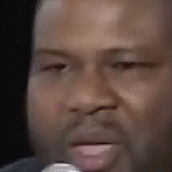}
            \end{subfigure}
            \begin{subfigure}{0.09\linewidth}
                \centering
                \includegraphics[width=\linewidth]{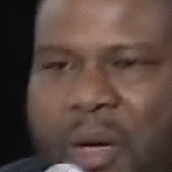}
            \end{subfigure}
            \begin{subfigure}{0.09\linewidth}
                \centering
                \includegraphics[width=\linewidth]{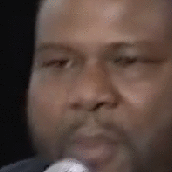}
            \end{subfigure}
            \begin{subfigure}{0.09\linewidth}
                \centering
                \includegraphics[width=\linewidth]{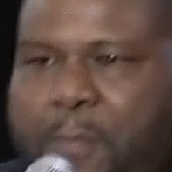}
            \end{subfigure}
            \begin{subfigure}{0.09\linewidth}
                \centering
                \includegraphics[width=\linewidth]{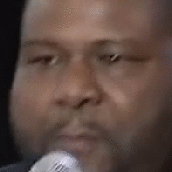}
            \end{subfigure}
            \begin{subfigure}{0.09\linewidth}
                \centering
                \includegraphics[width=\linewidth]{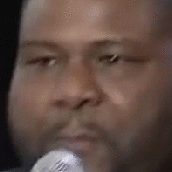}
            \end{subfigure}
            \begin{subfigure}{0.09\linewidth}
                \centering
                \includegraphics[width=\linewidth]{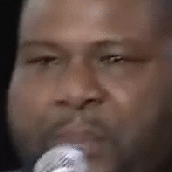}
            \end{subfigure}
            \begin{subfigure}{0.09\linewidth}
                \centering
                \includegraphics[width=\linewidth]{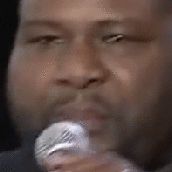}
            \end{subfigure}
            \begin{subfigure}{0.09\linewidth}
                \centering
                \includegraphics[width=\linewidth]{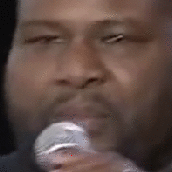}
            \end{subfigure}
            \begin{subfigure}{0.09\linewidth}
                \centering
                \includegraphics[width=\linewidth]{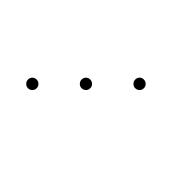}
            \end{subfigure}
        }
        \caption{Sample \texttt{6178} is labeled as a negative (\emph{not speaking}) sample by the automatic transformation from LRS3 to VVAD dataset. On the human accuracy level test 100\% of the subjects classified the sample as positive (\emph{speaking}) sample. Beat boxing is not considered speech in the LRS3 dataset, which causes the wrong label.}
        \label{fig:6178}
    \end{figure}
    \begin{figure}
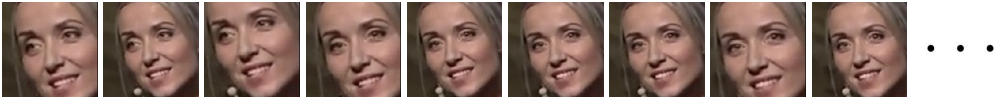

        \centering
        \parbox{\textwidth}{
            \begin{subfigure}{0.09\linewidth}
                \centering
                \includegraphics[width=\linewidth]{images/31366_faceImage_1/cropped_img-10.png}
            \end{subfigure}
            \begin{subfigure}{0.09\linewidth}
                \centering
                \includegraphics[width=\linewidth]{images/31366_faceImage_1/cropped_img-11.png}
            \end{subfigure}
            \begin{subfigure}{0.09\linewidth}
                \centering
                \includegraphics[width=\linewidth]{images/31366_faceImage_1/cropped_img-12.png}
            \end{subfigure}
            \begin{subfigure}{0.09\linewidth}
                \centering
                \includegraphics[width=\linewidth]{images/31366_faceImage_1/cropped_img-13.png}
            \end{subfigure}
            \begin{subfigure}{0.09\linewidth}
                \centering
                \includegraphics[width=\linewidth]{images/31366_faceImage_1/cropped_img-14.png}
            \end{subfigure}
            \begin{subfigure}{0.09\linewidth}
                \centering
                \includegraphics[width=\linewidth]{images/31366_faceImage_1/cropped_img-15.png}
            \end{subfigure}
            \begin{subfigure}{0.09\linewidth}
                \centering
                \includegraphics[width=\linewidth]{images/31366_faceImage_1/cropped_img-16.png}
            \end{subfigure}
            \begin{subfigure}{0.09\linewidth}
                \centering
                \includegraphics[width=\linewidth]{images/31366_faceImage_1/cropped_img-17.png}
            \end{subfigure}
            \begin{subfigure}{0.09\linewidth}
                \centering
                \includegraphics[width=\linewidth]{images/31366_faceImage_1/cropped_img-18.png}
            \end{subfigure}
            \begin{subfigure}{0.09\linewidth}
                \centering
                \includegraphics[width=\linewidth]{images/dots.jpg}
            \end{subfigure}
        }
        \caption{Sample \texttt{31366} is labeled as a positive (\emph{speaking}) sample by the automatic transformation from LRS3 to VVAD dataset. On the human accuracy level test 100\% of the subjects classified the sample as negative (\emph{not speaking}) sample. The fast movement of the head while producing only a small movement of the lips causes the wrong label.}
        \label{fig:31366}
    \end{figure}
\fi
\makeatother

\section{INITIAL EXPERIMENTAL RESULTS}

\textbf{Pre-trained Models}. To show that the dataset can be efficiently used to train a VVAD, we implemented and trained CNN-LSTM models with our dataset as baselines. As described earlier speech cannot be effectively classified with a single image, which motivates the use of recurrent neural networks.

We evaluate the use of LSTM cells, as described in \cite{HochSchm97}. We use standard architectures as a backbone, which are wrapped by a \emph{TimeDistributed} wrapper in order to transform them into a recurrent network that can process a sequence of images. \emph{TimeDistributed} is a wrapper provided by Keras \cite{chollet2015keras} which basically copies a model for all timesteps, to effectively handle time series and sequences.
The sequence can be processed by a LSTM Layer to make temporal sense, while the last Dense Layer is used to make the classification. Experiments have shown that a single Dense layer with 512 units on top of a LSTM layer with 32 units show good results. We use a $200 \times 200$ pixels input image size on one or two frames for initial testing.

We use DenseNet \cite{huang2018densely}, MobileNet \cite{howard2017mobilenets} and VGGFace \cite{Parkhi15} as backbone networks in the \emph{TimeDistributed} wrapper. These models are pre-trained and used as is from the \emph{keras-applications} library.

All models were trained using Stochastic Gradient Descent, with a starting learning rate $\alpha = 0.01$ and decaying as needed. Models were trained until convergence, which varied between 80 to 200 epochs. A binary cross-entropy loss is used, and each network has an output layer with a single neuron and a sigmoid activation. All architectures and hidden layers use a ReLU activation.

Our results are presented in Table \ref{tb:baselineAccuracies}. It shows that DenseNet, MobileNet and VGGFace improve by around 2.3\% using one more frame. Our results also shows that MobileNetV1 and DenseNet121 perform better than the corresponding model alteration. We will refer as MobileNet and DenseNet to MobileNetV1 and DenseNet121 respectively.

\textbf{End-to-End Learning}. In this section we evaluate end-to-end models trained from scratch, using not just face images but also other features such as lips and their features. Since evaluating for all 38 frames is not always possible (depending on access to GPUs with large amounts of RAM), only the MobileNet as the smallest of the base models is taken further into consideration. For this experiment we use $96 \times 96$ input image sizes for image features.

In comparison MobileNet contains approximately 4.2 million parameters while DenseNet requires around double the amount with 8 million parameters and VGGFace has over 50 million parameters. Knowing this the MobileNet is a good compromise between performance and size, because it is able to consider more timesteps, which in the end can lead to even higher accuracy.

For the face and lip images a \emph{TimeDistributed} MobileNet is used, while the approaches learning on the vector features (facial and lip features) we use a single LSTM layer with 32 units and a single Dense layer with 512 units. Training methodology is the same as pre-trained models.

Our results are presented in Table \ref{tb:allAccuracies} It shows that even with the substantially smaller face features a validation accuracy of 89.79\% can be reached which is still higher than the human accuracy level. With this end-to-end learning approach on face images we were able to reach a very high accuracy of 92\% on the test set. This is higher than the reported human-level accuracy on the same dataset.

One interesting remark from our results is that learning from image data, even if it is from scratch, seems to outperform the use of facial or lip features by approximately 3\%, which we believe makes sense since an image might contain additional information that the pure facial or lip features do not contain. This shows the importance of using visual models for this problem.

\begin{table}[tb]
    \caption{Validation accuracies for the different baseline models using only one or two frame in full resolution ($200 \times 200$ pixels). Increased accuracy highlights the importance of the temporal domain in VVAD.}
    \label{tb:baselineAccuracies}
    \centering
    \begin{tabular}{lll}
        \toprule
        \textbf{Baseline Model} & \textbf{1-Frame Accuracy} & \textbf{2-Frame Accuracy} \\
        \midrule
        DenseNet201             & 73.08 \%                  & -                         \\
        DenseNet121             & 73.17 \%                  & 75.34 \%                  \\
        MobileNetV2             & 67.45 \%                  & -                         \\
        MobileNetV1             & 69.56 \%                  & 72.11\%                   \\
        VGGFace                 & 71.96 \%                  & 74.36 \%                  \\
        \bottomrule
    \end{tabular}
\end{table}

\begin{figure}
    \centering
    \includegraphics[width=\figureWidth]{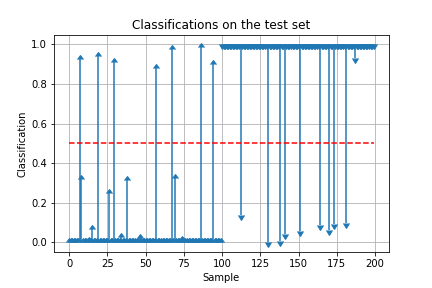}
    \caption{Visualization of predictions on the test set for MobileNet trained on face images. Each arrow represents the prediction confidence, with the first 100 samples being negative (not speaking), and the remaining 100 samples being positive (speaking).}
    \label{fig:testFaceImages}
\end{figure}

\textbf{Prediction Analysis}.  The classifications of the samples from the test set can be seen in Figure \ref{fig:testFaceImages}.
The first 100 samples are negative samples while the last 100 samples are positive ones and the arrows show the probability, given by the model, that this sample belongs to the positive class.
The red dashed line is the decision boundary on which the model decides its classifications.
This visualizes how certain the model is with its predictions. Many predictions are incorrect with a high confidence, indicating overconfidence, which also motivates the use of properly calibrated and Bayesian neural network models \cite{matin2020}

\makeatletter%
\if@twocolumn%
    \figureWidth=\columnwidth
\else% \@twocolumnfalse
    \figureWidth=0.6\columnwidth
\fi
\makeatother

\begin{table}[t]
    \caption{Overview of the validation performance of the different features on MobileNet using all 38 frames at $96 \times 96$ pixels}
    \label{tb:allAccuracies}
    \centering
    \begin{tabular}{lll}
        \toprule
        \textbf{Feature} & \textbf{Validation Accuracy} & \textbf{Test Accuracy} \\
        \midrule
        Face Images      & 94.05 \%                     & 92.0 \%                \\
        Lip Images       & 93.98 \%                     & 92.0 \%                \\
        Face Features    & 89.79 \%                     & 89.0 \%                \\
        Lip Features     & 89.93 \%                     & 89.0 \%                \\
        \midrule
        Human Level      & -                            & 87.93 \%               \\
        \bottomrule
    \end{tabular}
\end{table}

\section{CONCLUSIONS AND FUTURE WORK}
In this work we present the construction of the VVAD-LRS3 dataset using an automated pipeline to construct VVAD samples from LRS3 samples, we also show that these samples are not labeled perfectly, but they can still be used to learn a robust VVAD system. The VVAD-LRS3 dataset provides four kinds of features: facial and lip images, and facial and lip landmark features.

We provide baselines on our dataset using pre-trained and end-to-end neural network architectures on all feature kinds. Face images with end-to-end architectures seem to perform best with a validation accuracy of up to $94 \%$, while landmark features on face and lips seem to perform the worse at around $89 \%$ validation accuracy. We also show that up to 38 frames are required to obtain the highest predictive performance for this task.

Although the performance shows to be better than human accuracy and the presented solutions seem to be robust enough to handle outliers it may be possible to improve the results with a cleaned dataset. The cleaning can be done by manually testing all labels and correct or remove wrong labeled samples or by enhancing the algorithm to reduce the number of wrong labels.

Due to the comparability of the test results with the human accuracy level it was only possible to use the 200 randomly seeded samples used for the human accuracy test as the test set for the trained models, although it was described as best practice to hold back at least 10\% of the data for testing.
If the test set would be bigger and comparability to the human performance can be secured the test results would have an even stronger meaning than right now.
A larger amount of samples that were tested on humans would make it possible to examine the relationship between DNNs for VVAD and the human brains approach to VVAD more closely.
Furthermore it is hard to determine a ground truth for the data because human classification varied for some of the samples.
But in general the human classification and the data creation through the automatic pipeline have a significant similarity which allows us to use the data effectively as is.
Experiments with trained models in real human-robot interaction can hopefully be conducted in the future.
We hope that the community benefits from our dataset and is able to produce learning algorithms that can produce a robust VVAD system for social robots.

The dataset is publicly available under \href{https://www.kaggle.com/adrianlubitz/vvadlrs3}{https://www.kaggle.com/adrianlubitz/vvadlrs3}.
With a large scale publicly available dataset for VVAD the research on this topic can be massively accelerated.
Furthermore we were able to publish some of the trained models on PyPI \cite{PyPI} under  \href{https://pypi.org/project/vvadlrs3/}{https://pypi.org/project/vvadlrs3/} to make it easier to develop applications.

\clearpage
\bibliographystyle{plainnat}
\bibliography{references}

%%%%%%%%%%%%%%%%%%%%%%%%%%%%%%%%%%%%%%%%%%%%%%%%%%%%%%%%%%%%

\clearpage
\appendix

\section{Appendix}

% Include extra information in the appendix.
% This section will often be part of the supplemental material.
% Please see the call on the NeurIPS website for links to additional guides on dataset publication.

%%%%%%%%%%%%%%%%%%%%%%%%%%%%%%%%%%%%%%%%%%%%%%%%%%%
This submission includes the following supplementary materials:
\begin{enumerate}
    \item Link to dataset:  \href{https://www.kaggle.com/adrianlubitz/vvadlrs3}{https://www.kaggle.com/adrianlubitz/vvadlrs3}
    \item Hosting, licensing, and maintenance plan can be viewed under  \href{https://www.kaggle.com/adrianlubitz/vvadlrs3}{https://www.kaggle.com/adrianlubitz/vvadlrs3}.
    \item Dataset documentation using datasheets for datasets \cite{gebru_datasheets_2020}
\end{enumerate}
%%%%%%%%%%%%%%%%%%%%%%%%%%%%%%%%%%%%%%%%%%%%%%%%%%%%

\section{Datasheet for VVAD-LRS3 Dataset}

%%%%%%%%%%%%%%%%%%%%%%%%%%%%%%%%%%%%%%%%%%%%%%%%%%%%%%%%%%%%%%%%%%%%%%%%%%%%%%%%
\subsection{Motivation For Datasheet Creation}

\textcolor{blue}{\subsubsection{Why was the datasheet created? (e.g., was there a specific task in mind? was there a specific gap that needed to be filled?)}}

Automatic Speech Recognition (ASR) and Natural Language Processing (NLP) enable robots to have great understanding about what humans want to tell them.
But there are some social situations where hearing alone is not good enough to give a socially acceptable answer.
Robots need to know when they should start and stop to process audio streams to react only to speech directed towards them.
VVAD enables robots to detect if a human is directly speaking to it. This is very helpful in noisy environments where Voice Activity Detection alone is not sufficient.
Furthermore it can help to detect a speaker for situations where multiple people are in the robot's field of view.

\textcolor{blue}{\subsubsection{Has the dataset been used already? If so, where are the results so others can compare
                (e.g., links to published papers)?}}

The dataset has only been used for initial experiments described in the paper so far.

\textcolor{blue}{\subsubsection{What (other) tasks could the dataset be used for?}}

Outside the robotic context VVAD can be used in video conferencing systems to detect faulty or muted microphones (from visual input, if a person is speaking).
Furthermore it could manage to automatically mute and unmute microphones in videoconference to guarantee good voice quality for all participants.

\textcolor{blue}{\subsubsection{Who funded the creation dataset?}}

The creation of this dataset did not receive any funding. It was part of the Master Thesis of Adrian Lubitz at the University of Bremen.

The original LRS3 dataset was funded by various sources (UK EPSRC CDT in Autonomous Intelligent Machines and Systems, the Oxford-Google DeepMind Graduate Scholarship, and by the EPSRC Programme Grant Seebibyte EP/M013774/1).

\textcolor{blue}{\subsubsection{Any other comments?}}
N.A.

\subsection{Datasheet Composition}

\textcolor{blue}{\subsubsection{What are the instances?(that is, examples; e.g., documents, images, people, countries) Are there multiple types
                of instances? (e.g., movies, users, ratings; people, interactions between them; nodes, edges)}}
Short videos of human faces.

\textcolor{blue}{\subsubsection{How many instances are there in total (of each type, if appropriate)?}}
44,489 short videos of humans

\textcolor{blue}{\subsubsection{What data does each instance consist of ? “Raw”
                data (e.g., unprocessed text or images)? Features/attributes? Is there a label/target associated with
                instances? If the instances related to people, are subpopulations identified (e.g., by age, gender, etc.) and what is
                their distribution?}}
Each instance is available as
\begin{enumerate}
        \item series of images of faces
        \item series of images of lips
        \item series of 2D face features
        \item series of 2D lip features
\end{enumerate}

\textcolor{blue}{\subsubsection{Is there a label or target associated with each instance? If so, please
                provide a description.}}
The data has binary labels. True for \emph{speaking} and False for \emph{not speaking}.

\textcolor{blue}{\subsubsection{Is any information missing from individual instances? If so, please
                provide a description, explaining why this information is missing (e.g., because it was unavailable). This does not include intentionally removed
                information, but might include, e.g., redacted text.}}
There is no data missing but as a side note it can be mentioned that over 95\% of \emph{speaking} samples where rejected to keep the dataset ratio 1-to-1.

\textcolor{blue}{\subsubsection{Are relationships between individual instances made explicit (e.g.,
                users’ movie ratings, social network links)? If so, please describe
                how these relationships are made explicit.}}
N.A.

\textcolor{blue}{\subsubsection{Does the dataset contain all possible instances or is it a sample (not
                necessarily random) of instances from a larger set? If the dataset is
                a sample, then what is the larger set? Is the sample representative of the
                larger set (e.g., geographic coverage)? If so, please describe how this
                representativeness was validated/verified. If it is not representative of the
                larger set, please describe why not (e.g., to cover a more diverse range of
                instances, because instances were withheld or unavailable).}}

Over 95\% of \emph{speaking} samples in the original data where rejected to keep the dataset ratio 1-to-1 and keep the dataset size to a minimum.
Positive samples were randomly rejected to keep the diversity and initial experiments showed that the size of the dataset is sufficient to train a classifier that reaches superhuman performance.

\textcolor{blue}{\subsubsection{Are there recommended data splits (e.g., training, development/validation, testing)? If so, please provide a description of these
                splits, explaining the rationale behind them.}}
\begin{table}[t]
        \caption{Number of samples for training, validation and test splits of the VVAD-LRS3 dataset}
        \label{tb:samples}
        \centering
        \begin{tabular}{lll}
                \toprule
                \textbf{Training Set} & \textbf{Validation Set} & \textbf{Test Set} \\
                \midrule
                37646 Samples         & 6643 Samples            & 200 Samples       \\
                \bottomrule
        \end{tabular}

\end{table}
From the whole dataset approximately 15\% of the data was taken as validation set as it is described as best practice to take 10-15\% of the data as validation data.
To ensure comparability between the human accuracy test and the accuracy reached by the trained classifier the test data is the same data which was used for the human accuracy test.

\textcolor{blue}{\subsubsection{Are there any errors, sources of noise, or redundancies in the
                dataset? If so, please provide a description.}}

The transcript of the talk for each instance is used to decide if the human is speaking or not, so any error in the transcript could produce erroneous labels.

There are some cases of ambiguities, for example a case of beatboxing where vocal sound is produced with minimum lip movement, and could be labeled as not speaking.

\textcolor{blue}{\subsubsection{Is the dataset self-contained, or does it link to or otherwise rely on
                external resources (e.g., websites, tweets, other datasets)? If it links
                to or relies on external resources, a) are there guarantees that they will
                exist, and remain constant, over time; b) are there official archival versions
                of the complete dataset (i.e., including the external resources as they existed at the time the dataset was created); c) are there any restrictions
                (e.g., licenses, fees) associated with any of the external resources that
                might apply to a future user? Please provide descriptions of all external
                resources and any restrictions associated with them, as well as links or
                other access points, as appropriate.}}
The dataset is self-contained.

\textcolor{blue}{\subsubsection{Any other comments?}}
N.A.

\subsection{Collection Process}

\textcolor{blue}{\subsubsection{What mechanisms or procedures were used to collect the data (e.g.,
                hardware apparatus or sensor, manual human curation, software program, software API)? How were these mechanisms or procedures validated?}}
The data is extracted from the LRS3 dataset by an automatic pipeline.
Samples were manually validated on a random selection.

\textcolor{blue}{\subsubsection{How was the data associated with each instance acquired? Was the
                data directly observable (e.g., raw text, movie ratings), reported by subjects (e.g., survey responses), or indirectly inferred/derived from other data
                (e.g., part-of-speech tags, model-based guesses for age or language)?
                If data was reported by subjects or indirectly inferred/derived from other
                data, was the data validated/verified? If so, please describe how.}}
The data derived from the LRS3 dataset and samples were manually validated on a random selection.

\textcolor{blue}{\subsubsection{If the dataset is a sample from a larger set, what was the sampling strategy (e.g., deterministic, probabilistic with specific sampling probabilities)?}}
Over 95\% of positive samples were randomly rejected to keep the diversity and initial experiments showed that the size of the dataset is sufficient to train a classifier that reaches superhuman performance.

\textcolor{blue}{\subsubsection{Who was involved in the data collection process (e.g., students,
                crowdworkers, contractors) and how were they compensated (e.g.,
                how much were crowdworkers paid)?}}
N.A.

\textcolor{blue}{\subsubsection{Over what timeframe was the data collected? Does this timeframe
                match the creation timeframe of the data associated with the instances
                (e.g., recent crawl of old news articles)? If not, please describe the timeframe in which the data associated with the instances was created.}}
N.A.

\subsection{Data Preprocessing}

\textcolor{blue}{\subsubsection{Was any preprocessing/cleaning/labeling of the data done (e.g., discretization or bucketing, tokenization, part-of-speech tagging, SIFT
                feature extraction, removal of instances, processing of missing values)? If so, please provide a description. If not, you may skip the remainder of the questions in this section.}}
There is only minimal pre-processing, normalization of the data for training a model, all specified in our paper.

\textcolor{blue}{\subsubsection{Was the “raw” data saved in addition to the preprocessed/cleaned/labeled data (e.g., to support unanticipated
                future uses)? If so, please provide a link or other access point to the
                “raw” data.}}
N.A.
\textcolor{blue}{\subsubsection{Is the software used to preprocess/clean/label the instances available? If so, please provide a link or other access point.}}
\href{https://github.com/adrianlubitz/VVAD}{https://github.com/adrianlubitz/VVAD} - More instructions on how to use the data creation pipeline will follow.

\textcolor{blue}{\subsubsection{Does this dataset collection/processing procedure
                achieve the motivation for creating the dataset
                stated in the first section of this datasheet? If not,
                what are the limitations?}}
Yes. First experiments show good results. One limitation is that many images in a sequence are required to achieve good accuracy. A small number of frames (< 10) will decrease accuracy to around 75-80\%.

\textcolor{blue}{\subsubsection{Any other comments}}
N.A.

\subsection{Dataset Distribution}

\textcolor{blue}{\subsubsection{How will the dataset be distributed? (e.g., tarball on
                website, API, GitHub; does the data have a DOI and is it
                archived redundantly?)}}
The dataset is already publicly available and can be found under \href{https://www.kaggle.com/adrianlubitz/vvadlrs3}{https://www.kaggle.com/adrianlubitz/vvadlrs3} and has the DOI:\texttt{ 10.34740/kaggle/dsv/2236600}

\textcolor{blue}{\subsubsection{When will the dataset be released/first distributed?
                What license (if any) is it distributed under?}}
The dataset is already publicly available in Kaggle.

It is released and distributed under the LGPLv2 with the following additional note:

\emph{Note that the license for the iBUG 300-W dataset which was used for face and lip features excludes commercial use.}

The original data (YouTube TED talks) is available under the \href{https://www.ted.com/about/our-organization/our-policies-terms/ted-com-terms-of-use}{TED Terms of Use} and the \href{https://creativecommons.org/licenses/by-nc-nd/4.0/legalcode}{Creative Commons BY-NC-ND 4.0 license}, according to the information on the \href{https://www.robots.ox.ac.uk/~vgg/data/lip_reading/lrs3.html}{LRS3 dataset}.

\textcolor{blue}{\subsubsection{Are there any copyrights on the data?}}
Yes.

Copyright of the YouTube TED Talks is owned by TED, as well as transcripts.

\textcolor{blue}{\subsubsection{Are there any fees or access/export restrictions?}}
No.

\textcolor{blue}{\subsubsection{Any other comments?}}
N.A.

\subsection{Dataset Maintenance}

\textcolor{blue}{\subsubsection{Who is supporting/hosting/maintaining the
                dataset?}}
Adrian Lubitz. All comments/concerns can be sent to adrianlubitz@gmail.com

\textcolor{blue}{\subsubsection{Will the dataset be updated? If so, how often and
                by whom?}}
The dataset is as is. We will only update if there is a good reason to do so, like incorrect labels or ethical issues.

\textcolor{blue}{\subsubsection{How will updates be communicated? (e.g., mailing
                list, GitHub)}}
Kaggle Version Tag

\textcolor{blue}{\subsubsection{If the dataset becomes obsolete how will this be
                communicated?}}
In the Kaggle dataset description, and possible updating this paper (if published) on arXiv.

\textcolor{blue}{\subsubsection{Is there a repository to link to any/all papers/systems that use this dataset?}}
No usage by now. Usage will be documented in the GitHub Project and the Kaggle Site.

\textcolor{blue}{\subsubsection{If others want to extend/augment/build on this
                dataset, is there a mechanism for them to do so?
                If so, is there a process for tracking/assessing the
                quality of those contributions. What is the process
                for communicating/distributing these contributions
                to users?}}
Nothing in place so far.

\subsection{Legal and Ethical Considerations}

\textcolor{blue}{\subsubsection{Were any ethical review processes conducted (e.g., by an institutional review board)? If so, please provide a description of these review
                processes, including the outcomes, as well as a link or other access point
                to any supporting documentation.}}
No. As far as we are aware, it was not required.

\textcolor{blue}{\subsubsection{Does the dataset contain data that might be considered confidential
                (e.g., data that is protected by legal privilege or by doctor patient confidentiality, data that includes the content of individuals non-public
                communications)? If so, please provide a description.}}
No.

\textcolor{blue}{\subsubsection{Does the dataset contain data that, if viewed directly, might be offensive, insulting, threatening, or might otherwise cause anxiety? If so,
                please describe why}}
No.

\textcolor{blue}{\subsubsection{Does the dataset relate to people? If not, you may skip the remaining
                questions in this section.}}
Yes.

\textcolor{blue}{\subsubsection{Does the dataset identify any subpopulations (e.g., by age, gender)?
                If so, please describe how these subpopulations are identified and provide
                a description of their respective distributions within the dataset.}}
No.

\textcolor{blue}{\subsubsection{Is it possible to identify individuals (i.e., one or more natural persons), either directly or indirectly (i.e., in combination with other
                data) from the dataset? If so, please describe how.}}
Yes. The dataset contains facial images which can identify a person.

\textcolor{blue}{\subsubsection{Does the dataset contain data that might be considered sensitive in
                any way (e.g., data that reveals racial or ethnic origins, sexual orientations, religious beliefs, political opinions or union memberships, or
                locations; financial or health data; biometric or genetic data; forms of
                government identification, such as social security numbers; criminal
                history)? If so, please provide a description.}}
No.

\textcolor{blue}{\subsubsection{Did you collect the data from the individuals in question directly, or
                obtain it via third parties or other sources (e.g., websites)?}}
The data is obtained from the TED Talks on YouTube, so it would be collected by a third party.

\textcolor{blue}{\subsubsection{Were the individuals in question notified about the data collection?
                If so, please describe (or show with screenshots or other information) how
                notice was provided, and provide a link or other access point to, or otherwise reproduce, the exact language of the notification itself.}}
Yes, speakers have to sign a standard release for TED to record their talk and publish it in the internet, so individuals are fully aware.

\textcolor{blue}{\subsubsection{Did the individuals in question consent to the collection and use of
                their data? If so, please describe (or show with screenshots or other
                information) how consent was requested and provided, and provide a link
                or other access point to, or otherwise reproduce, the exact language to
                which the individuals consented.}}
Yes.

\textcolor{blue}{\subsubsection{If consent was obtained, were the consenting individuals provided
                with a mechanism to revoke their consent in the future or for certain
                uses? If so, please provide a description, as well as a link or other access
                point to the mechanism (if appropriate).}}
We are not aware of mechanisms for revoking consent, this would depend on mechanisms at TED.

\textcolor{blue}{\subsubsection{Has an analysis of the potential impact of the dataset and its use
                on data subjects (e.g., a data protection impact analysis)been conducted? If so, please provide a description of this analysis, including the
                outcomes, as well as a link or other access point to any supporting documentation.}}
No.

\textcolor{blue}{\subsubsection{Any other comments?}}
N.A.

\end{document}